\documentclass{article}
\pdfoutput=1
\PassOptionsToPackage{numbers}{natbib}

\usepackage[final]{nips_2016}

\usepackage[utf8]{inputenc} 
\usepackage[T1]{fontenc}    
\usepackage{hyperref}       
\usepackage{url}            
\usepackage{booktabs}       
\usepackage{amsfonts}       
\usepackage{nicefrac}       
\usepackage{microtype}      
\usepackage{amsmath,amsfonts}
\usepackage{mathtools}
\usepackage{bbm} 
\usepackage{todonotes}
\usepackage{multirow}

\newtheorem{theorem}{Theorem}[section]

\newtheorem{lemma}[theorem]{Lemma}

\setcitestyle{square}

\providecommand{\openbox}{\leavevmode
  \hbox to.77778em{%
  \hfil\vrule
  \vbox to.675em{\hrule width.6em\vfil\hrule}%
  \vrule\hfil}}
\makeatletter
\DeclareRobustCommand{\qed}{%
  \ifmmode
    \eqno \def\@badmath{$$}
    \let\eqno\relax \let\leqno\relax \let\veqno\relax
    \hbox{\openbox}%
  \else
    \leavevmode\unskip\penalty9999 \hbox{}\nobreak\hfill
    \quad\hbox{\openbox}%
  \fi
}
\makeatother

\title{Full-Capacity Unitary Recurrent Neural Networks}

\author{
  Scott Wisdom$^{1}$\thanks{Equal contribution}\hphantom{$^*$}, Thomas Powers$^{1}$$^*$, John R. Hershey$^2$, Jonathan Le Roux$^2$, and Les Atlas$^1$
  \\
  $^1$ Department of Electrical Engineering, University of Washington \\
  \texttt{\{swisdom, tcpowers, atlas\}@uw.edu} \\
  $^2$ Mitsubishi Electric Research Laboratories (MERL) \\
  \texttt{\{hershey, leroux\}@merl.com}
}

\begin{document}


\maketitle


\begin{abstract}
  Recurrent neural networks are powerful models for processing sequential data, but they are generally plagued by vanishing and exploding gradient problems. Unitary recurrent neural networks (uRNNs), which use unitary recurrence matrices, have recently been proposed as a means to avoid these issues. However, in previous experiments, the recurrence matrices were restricted to be a product of parameterized unitary matrices, and 
  an open question remains: 
  when does such a parameterization fail to represent all unitary matrices, and how does this restricted representational capacity limit what can be learned?
  To address this question,
  we propose full-capacity uRNNs that optimize their recurrence matrix over all unitary matrices, leading to significantly improved performance over uRNNs that use a restricted-capacity recurrence matrix. 
  Our contribution consists of two main components. 
  First, we provide a theoretical argument to determine if a unitary parameterization has restricted capacity. Using this argument, we show that a recently proposed unitary parameterization has restricted capacity for hidden state dimension greater than 7.
  Second, we show how a complete, full-capacity unitary recurrence matrix can be optimized over the differentiable manifold of unitary matrices. The resulting multiplicative gradient step is very simple and does not require gradient clipping or learning rate adaptation.
  We confirm the utility of our claims by empirically evaluating our new full-capacity uRNNs on both synthetic and natural data, achieving superior performance compared to both LSTMs and the original restricted-capacity uRNNs.
\end{abstract}

\section{Introduction}

Deep feed-forward and recurrent neural networks have been shown to be remarkably effective in a wide variety of problems.   A primary difficulty in training using gradient-based methods has been the so-called  \emph{vanishing or exploding gradient} problem, in which the instability of the gradients over multiple layers can impede learning  \cite{bengio_learning_1994,hochreiter_gradient_2001}.   This problem is particularly keen for recurrent networks, since the repeated use of the recurrent weight matrix can magnify any instability. 

This problem has been addressed in the past by various means, including gradient clipping \cite{pascanu_difficulty_2012}, using orthogonal matrices for initialization of the recurrence matrix \cite{saxe_exact_2013,le_simple_2015}, or by using pioneering architectures such as long short-term memory (LSTM) recurrent networks  \cite{hochreiter_long_1997} or gated recurrent units~\cite{cho_properties_2014}.  Recently, several innovative architectures have been introduced to improve information flow in a network:  residual networks, which directly pass information from previous layers up in a feed-forward network   \cite{he_deep_2015}, and attention networks, which allow a recurrent network to access past activations \cite{mnih_recurrent_2014}. 
The idea of using a unitary recurrent weight matrix was introduced so that the gradients are inherently stable and do not vanish or explode \cite{arjovsky_unitary_2016}. The resulting unitary recurrent neural network (uRNN) is complex-valued and uses a complex form of the rectified linear activation function. However, this idea was investigated using, as we show, a potentially restricted form of unitary matrices.


The two main components of our contribution can be summarized as follows: 

1) We provide a theoretical argument to determine the smallest dimension $N$ for which any parameterization of the unitary recurrence matrix does not cover the entire set of all unitary matrices. The argument relies on counting real-valued parameters and using Sard's theorem to show that the smooth map from these parameters to the unitary manifold is not onto.
Thus, we can show that a previously proposed parameterization \cite{arjovsky_unitary_2016} cannot represent all unitary matrices larger than $7\times 7$. Thus, such a parameterization results in what we refer to as a {\it restricted-capacity} unitary recurrence matrix.

2) To overcome the limitations of restricted-capacity parameterizations, we propose a new method for stochastic gradient descent for training the unitary recurrence matrix, which constrains the gradient to lie on the differentiable manifold of unitary matrices. This approach allows us to directly optimize a complete, or {\it full-capacity}, unitary matrix.
Neither restricted-capacity nor full-capacity unitary matrix optimization require gradient clipping. Furthermore, full-capacity optimization still achieves good results without adaptation of the learning rate during training.

To test the limitations of a restricted-capacity representation and to confirm that our full-capacity uRNN does have practical implications, we test restricted-capacity and full-capacity uRNNs on both synthetic and natural data tasks. These tasks include synthetic system identification, long-term memorization, frame-to-frame prediction of speech spectra, and pixel-by-pixel classification of handwritten digits.
Our proposed full-capacity uRNNs generally achieve equivalent or superior performance on synthetic and natural data compared to both LSTMs \cite{hochreiter_long_1997} and the original restricted-capacity uRNNs \cite{arjovsky_unitary_2016}.



In the next section, we give an overview of unitary recurrent neural networks.
Section 3 presents our first contribution: the theoretical argument to determine if any unitary parameterization has restricted-capacity.
Section 4 describes our second contribution, where we show how to optimize a full-capacity unitary matrix. We confirm our results with simulated and natural data in Section 5 and present our conclusions in Section 6.

\section{Unitary recurrent neural networks}

The uRNN proposed by Arjovsky et al.~\cite{arjovsky_unitary_2016} consists of the following nonlinear dynamical system that has real- or complex-valued
inputs 
${\bf x}_t$ of dimension $M$, 
complex-valued hidden states
${\bf h}_t$ of dimension $N$, 
and real- or complex-valued outputs
${\bf y}_t$ of dimension $L$:
\begin{equation}
\label{eq:dynsys}
\begin{aligned}
    {\bf h}_t
    =&
    \sigma_{\bf b}
    \left(
        {\bf W}{\bf h}_{t-1}+{\bf V}{\bf x}_t
    \right) \\
    {\bf y}_t=&{\bf U}{\bf h}_t+{\bf c},
\end{aligned}
\end{equation}
where ${\bf y}_t=\mathrm{Re}\{{\bf U}{\bf h}_t+{\bf c}\}$ if the outputs ${\bf y}_t$ are real-valued. The element-wise nonlinearity $\sigma$ is
\begin{equation}
    \label{eq:nonlin}
    \left[\sigma_{\bf b}({\bf z})\right]_i=
    \begin{cases}
        (|z_i|+b_i)\frac{z_i}{|z_i|},
        &\mbox{if } |z_i|+b_i > 0, \\
    0,
    &\mbox{otherwise.}
    \end{cases}
\end{equation}
Note that this non-linearity consists in a soft-thresholding of the magnitude using the bias vector ${\bf b}$. Hard-thresholding would set the output of $\sigma$ to $z_i$ if $|z_i|+b_i>0$.
The parameters of the uRNN are as follows:
    ${\bf W}\in U(N)$, unitary hidden state transition matrix; ${\bf V}\in \mathbb{C}^{N\times M}$, input-to-hidden transformation; ${\bf b}\in \mathbb{R}^{N}$, nonlinearity bias; ${\bf U}\in \mathbb{C}^{L\times N}$, hidden-to-output transformation; and ${\bf c}\in \mathbb{C}^{L}$, output bias.

 Arjovsky et al.~\cite{arjovsky_unitary_2016} propose the following parameterization of the unitary matrix ${\bf W}$:
\begin{equation}
    \label{eq:adhoc}
    {\bf W}_{u}(\theta_{u})
    =
    {\bf D}_3
    {\bf R}_2
    \mathcal{F}^{-1}
    {\bf D}_2
    {\bf P}
    {\bf R}_1
    \mathcal{F}
    {\bf D}_1,
\end{equation}
where ${\bf D}$ are diagonal unitary matrices, ${\bf R}$ are Householder reflection matrices \cite{householder_unitary_1958}, $\mathcal{F}$ is a discrete Fourier transform (DFT) matrix, and ${\bf P}$ is a permutation matrix. The resulting matrix ${\bf W}_{u}$ is unitary because all its component matrices are unitary. This decomposition is efficient because diagonal, reflection, and permutation matrices are $\mathcal{O}(N)$ to compute, and DFTs can be computed efficiently in $\mathcal{O}(N\log N)$ time using the fast Fourier transform (FFT). The parameter vector $\theta_u$ consists of $7N$ real-valued parameters: $N$ parameters for each of the $3$ diagonal matrices where ${ D}_{i,i}=e^{j\theta_i}$ and $2N$ parameters for each of the $2$ Householder reflection matrices, which are real and imaginary values of the complex reflection vectors ${\bf u}_i$: ${\bf R}_i={\bf I}-2\frac{{\bf u}_i{\bf u}_i^H}{\langle {\bf u}_i,{\bf u}_i \rangle}$.

\section{Estimating the representation capacity of structured unitary matrices}
\label{sec:capacity}

    In this section, 
    we state and prove a theorem
    that can be used to determine when any particular unitary parameterization does not have capacity to represent 
    all unitary matrices.
    As an application of this theorem, we show that the parameterization (\ref{eq:adhoc}) does not have the capacity to cover all $N\times N$ unitary matrices for $N>7$.
    First, we establish an upper bound on the number of real-valued parameters required to represent any $N\times N$ unitary matrix. Then, we state and prove our theorem.

    \begin{lemma}
        \label{lem:Nsq}
        The set of all unitary matrices is a manifold of dimension $N^2$.
    \end{lemma}
    \noindent {\bf Proof:}
    The set of all unitary matrices is the well-known unitary Lie group $U(N)$ \cite[\S 3.4]{gilmore2008lie}. A Lie group identifies group elements with points on a differentiable manifold \cite[\S 2.2]{gilmore2008lie}. The dimension of the manifold is equal to the dimension of the Lie algebra $\mathfrak{u}$, which is a vector space that is the tangent space at the identity element \cite[\S 4.5]{gilmore2008lie}. 
    For $U(N)$, the Lie algebra consists of all skew-Hermitian 
    matrices ${\bf A}$ \cite[\S 5.4]{gilmore2008lie}.
    A skew-Hermitian matrix is any ${\bf A}\in\mathbb{C}^{N\times N}$ such that ${\bf A}=-{\bf A}^H$, where $(\cdot)^H$ is the conjugate transpose. To determine the dimension of $U(N)$, we can determine the dimension of $\mathfrak{u}$. Because of the skew-Hermitian constraint, the diagonal elements of ${\bf A}$ are purely imaginary, which corresponds to $N$ real-valued parameters. Also, since $A_{i,j}=-A_{j,i}^*$, the upper and lower triangular parts of ${\bf A}$ are parameterized by $\frac{N(N-1)}{2}$ complex numbers, which corresponds to an additional $N^2-N$ real parameters. Thus, $U(N)$ is a manifold of dimension $N^2$. \qed
    
    \begin{theorem}
    \label{thm:sard}
    If a family of $N\times N$ unitary matrices is parameterized by $P$ real-valued parameters for $P<N^2$, then it cannot contain all $N\times N$ unitary matrices.
    \end{theorem}
    \noindent {\bf Proof:}
    We consider a family of unitary matrices that is parameterized by $P$ real-valued parameters through a smooth map $g:\mathcal{P}(P)\rightarrow\mathcal{U}(N^2)$ from the space of parameters $\mathcal{P}(P)$ to the space of all unitary matrices $\mathcal{U}(N^2)$. The space $\mathcal{P}(P)$ of parameters is considered as a $P$-dimensional manifold, while the space $\mathcal{U}(N^2)$ of all unitary matrices is an $N^2$-dimensional manifold according to lemma \ref{lem:Nsq}.
    Then, if $P<N^2$, Sard's theorem \cite{sard_measure_1942} implies that the image $g(\mathcal{P})$ of $g$ is of measure zero in $\mathcal{U}(N^2)$, and in particular $g$ is not onto. Since $g$ is not onto, there must exist a unitary matrix ${\bf W}\in\mathcal{U}(N^2)$ for which there is no corresponding input ${\bf P}\in\mathcal{P}(P)$ such that ${\bf W} = g({\bf P})$. Thus, if $P$ is such that $P<N^2$, the manifold $\mathcal{P}(P)$ cannot represent all unitary matrices in $\mathcal{U}(N^2)$. 
    \qed
    
    We now apply Theorem \ref{thm:sard} to the parameterization (\ref{eq:adhoc}). Note that the parameterization (\ref{eq:adhoc}) has $P=7N$ real-valued parameters. If we solve for $N$ in $7N<N^2$, we get $N>7$. Thus, the parameterization (\ref{eq:adhoc}) cannot represent all unitary matrices for dimension $N>7$. 

    \section{Optimizing full-capacity unitary matrices on the Stiefel manifold}
    \label{ssec:stiefel}
    
    In this section, we show how to get around the limitations of restricted-capacity parameterizations and directly optimize a full-capacity unitary matrix. We consider the Stiefel manifold of all $N\times N$ complex-valued matrices whose columns are $N$ orthonormal vectors in $\mathbb{C}^N$ \cite{tagare_notes_2011}. Mathematically, the Stiefel manifold is defined as
    \begin{equation}
        \mathcal{V}_N(\mathbb{C}^N)
        =
        \left\{
            {\bf W}\in\mathbb{C}^{N\times N} : {\bf W}^H{\bf W}={\bf I}_{N\times N}
        \right\}.
    \end{equation}
    For any ${\bf W}\in\mathcal{V}_N(\mathbb{C}^N)$, any matrix ${\bf Z}$ in the tangent space $\mathcal{T}_{\bf W}\mathcal{V}_N(\mathbb{C}^N)$ of the Stiefel manifold satisfies ${\bf Z}^H{\bf W}-{\bf W}^H{\bf Z}=0$ \cite{tagare_notes_2011}. 
    The Stiefel manifold becomes a Riemannian manifold when its tangent space is equipped with an inner product. Tagare \cite{tagare_notes_2011} suggests using the canonical inner product, given by 
    \begin{equation}
        \langle {\bf Z}_1, {\bf Z}_2 \rangle_c
        =
        \mathrm{tr}
        \left( 
            {\bf Z}_1^H({\bf I}-\frac{1}{2}{\bf W}{\bf W}^H){\bf Z}_2
        \right).
    \end{equation}
    Under this canonical inner product on the tangent space, the gradient in the Stiefel manifold of the loss function $f$ with respect to the matrix ${\bf W}$ is ${\bf A}{\bf W}$, where
    $
        {\bf A}={\bf G}^H{\bf W}-{\bf W}^H{\bf G}
    $
    is a skew-Hermitian matrix and ${\bf G}$ with ${G}_{i,j}=\frac{\delta f}{\delta W_{i,j}}$ is the usual gradient of the loss function $f$ with respect to the matrix ${\bf W}$ \cite{tagare_notes_2011}. Using these facts, Tagare \cite{tagare_notes_2011} suggests a descent curve along the Stiefel manifold at training iteration $k$ given by the matrix product of the Cayley transformation
    of ${\bf A}^{(k)}$ with the current solution ${\bf W}^{(k)}$:
    \begin{equation}
        \label{eq:stiefel_step}
        {\bf Y}^{(k)}(\lambda)
        =
        \left(
            {\bf I} + \frac{\lambda}{2}{\bf A}^{(k)}
        \right)^{-1}
        \left(
            {\bf I} - \frac{\lambda}{2}{\bf A}^{(k)}
        \right)
        {\bf W}^{(k)},
    \end{equation}
    where $\lambda$ is a learning rate and ${\bf A}^{(k)}={{\bf G}^{(k)}}^H{\bf W}^{(k)}-{{\bf W}^{(k)}}^H{\bf G}^{(k)}$. Gradient descent proceeds by performing updates 
    ${\bf W}^{(k+1)}={\bf Y}^{(k)}(\lambda).$
    Tagare \cite{tagare_notes_2011} suggests an Armijo-Wolfe search along the curve to adapt $\lambda$, but such a procedure would be expensive for neural network optimization since it requires multiple evaluations of the forward model and gradients. We found that simply using a fixed learning rate $\lambda$ often works well. Also, RMSprop-style scaling of the gradient ${\bf G}^{(k)}$ by a running average of the previous gradients' norms \cite{tieleman_lecture_2012} before applying the multiplicative step (\ref{eq:stiefel_step}) can improve convergence. The only additional substantial computation required beyond the forward and backward passes of the network is the $N\times N$ matrix inverse in (\ref{eq:stiefel_step}).

\section{Experiments}

    All models are implemented in Theano \cite{theano_development_team_theano:_2016}, based on the implementation of restricted-capacity uRNNs by \cite{arjovsky_unitary_2016}, available from \url{https://github.com/amarshah/complex_RNN}. All code to replicate our results is available from \url{https://github.com/stwisdom/urnn}. All models use RMSprop \cite{tieleman_lecture_2012} for optimization, except that full-capacity uRNNs optimize their recurrence matrices with a fixed learning rate using the update step (\ref{eq:stiefel_step}) and optional RMSprop-style gradient normalization.

    \subsection{Synthetic data}
    
    First, we compare the performance of full-capacity uRNNs to restricted-capacity uRNNs and LSTMs on two tasks  with synthetic data. The first task is synthetic system identification, where a uRNN must learn the dynamics of a target uRNN given only samples of the target uRNN's inputs and outputs. The second task is the copy memory problem, in which the network must recall a sequence of data after a long period of time. 
    
    \subsubsection{System identification}
    
    For the task of system identification, we consider the problem of learning the dynamics of a nonlinear dynamical system that has the form (\ref{eq:dynsys}), given a dataset of inputs and outputs of the system. We will draw a true system ${\bf W}_{sys}$ randomly from either a constrained set $\mathcal{W}_{u}$ of restricted-capacity unitary matrices using the parameterization ${\bf W}_u(\theta_u)$ in (\ref{eq:adhoc}) or from a wider set $\mathcal{W}_{g}$ of restricted-capacity unitary matrices that are guaranteed to lie outside $\mathcal{W}_u$. We sample from $\mathcal{W}_{g}$ by taking a matrix product of two unitary matrices drawn from $\mathcal{W}_{u}$.
    
    We use a sequence length of $T=150$, and we set the input dimension $M$ and output dimension $L$ both equal to the hidden state dimension $N$. The input-to-hidden transformation ${\bf V}$ and output-to-hidden transformation ${\bf U}$ are both set to identity, the output bias ${\bf c}$ is set to ${\bf 0}$, the initial state is set to ${\bf 0}$, and the hidden bias ${\bf b}$ is drawn from a uniform distribution in the range $[-0.11,-0.09]$. The hidden bias has a mean of $-0.1$ to ensure stability of the system outputs. Inputs are generated by sampling $T$-length i.i.d.\ sequences of zero-mean, diagonal and unit covariance circular complex-valued Gaussians of dimension $N$. The outputs are created by running the system (\ref{eq:dynsys}) forward on the inputs.
    
    We compare a restricted-capacity uRNN using the parameterization from (\ref{eq:adhoc}) and a full-capacity uRNN using Stiefel manifold optimization with no gradient normalization as described in Section \ref{ssec:stiefel}. We choose hidden state dimensions $N$ to test critical points predicted by our arguments in Section \ref{sec:capacity} of ${\bf W}_u(\theta_u)$ in (\ref{eq:adhoc}):
    $N\in\{4,6,7,8,16\}$. These dimensions are chosen to test below, at, and above the critical dimension of $7$.
    
    For all experiments, the number of training, validation, and test sequences are $20000$, $1000$, and $1000$, respectively. 
    Mean-squared error (MSE) is used as the loss function. The learning rate is $0.001$ with a batch size of $50$ for all experiments. 
    Both models use the same matrix drawn from $\mathcal{W}_u$ as initialization. To isolate the effect of unitary recurrence matrix capacity, we only optimize ${\bf W}$, setting all other parameters to true oracle values. For each method, we report the best test loss over $100$ epochs and over $6$ random initializations for the optimization.
    
    The results are shown in Table \ref{tab:synthgen}. ``${\bf W}_{sys}$ init.'' refers to the initialization of the true system unitary matrix ${\bf W}_{sys}$, which is sampled from either the restricted-capacity set $\mathcal{W}_u$ or the wider set $\mathcal{W}_g$.
    
    \begin{table}[h]
    \caption{Results for system identification in terms of best normalized MSE. $\mathcal{W}_u$ is the set of restricted-capacity unitary matrices from (\ref{eq:adhoc}), and $\mathcal{W}_g$ is a wider set of unitary matrices.}
      \label{tab:synthgen}
      \centering
      \begin{tabular}{ccccccc}
            \toprule
            ${\bf W}_{sys}$ init.\ & Capacity & $N=4$ & $N=6$ & $N=7$ & $N=8$ & $N=16$ \\
            \midrule
            $\mathcal{W}_u$
            &
            Restricted
            &
            $4.81\text{e}{-1}$
            &
            $\mathbf{6.75\text{e}{-3}}$
            &
            $3.53\text{e}{-1}$
            &
            $3.51\text{e}{-1}$
            &
            $7.30\text{e}{-1}$
            \\
            $\mathcal{W}_u$
            &
            Full
            &
            $\mathbf{1.28\text{e}{-1}}$
            &
            $3.03\text{e}{-1}$
            &
            $\mathbf{2.16\text{e}{-1}}$
            &
            $\mathbf{5.04\text{e}{-2}}$
            &
            $\mathbf{1.28\text{e}{-1}}$
            \\
            \midrule
            $\mathcal{W}_g$
            &
            Restricted
            &
            $\mathbf{3.21\text{e}{-4}}$
            &
            $\mathbf{3.36\text{e}{-1}}$
            &
            $3.36\text{e}{-1}$
            &
            $2.69\text{e}{-1}$
            &
            $7.60\text{e}{-1}$
            \\
            $\mathcal{W}_g$
            &
            Full
            &
            $8.72\text{e}{-2}$
            &
            $3.86\text{e}{-1}$
            &
            $\mathbf{2.62\text{e}{-1}}$
            &
            $\mathbf{7.22\text{e}{-2}}$
            &
            $\mathbf{1.00\text{e}{-6}}$
            \\
            \bottomrule
          \end{tabular}
    \end{table}
    Notice that for $N<7$, the restricted-capacity uRNN achieves comparable or better performance than the full-capacity uRNN. At $N=7$, the restricted-capacity and full-capacity uRNNs achieve relatively comparable performance, with the full-capacity uRNN achieving slightly lower error. For $N>7$, the full-capacity uRNN always achieves better performance versus the restricted-capacity uRNN. This result confirms our theoretical arguments that the restricted-capacity parameterization in (\ref{eq:adhoc}) lacks the capacity to model all matrices in the unitary group for $N>7$ and indicates the advantage of using a full-capacity unitary recurrence matrix.
    
    \subsubsection{Copy memory problem}
    
    The experimental setup follows the copy memory problem from \cite{arjovsky_unitary_2016}, which itself was based on the experiment from  \cite{hochreiter_long_1997}. We consider alternative hidden state dimensions and extend the sequence lengths to $T=1000$ and $T=2000$, which are longer than the maximum length of $T=750$ considered in previous literature. 
    
    In this task, the data is a vector of length $T+20$ and consists of elements from 10 categories. The vector begins with a sequence of 10 symbols sampled uniformly from categories 1 to 8. The next $T-1$ elements of the vector are the ninth 'blank' category, followed by an element from the tenth category, the `delimiter'. The remaining ten elements are `blank'. 
    The task is to output $T+10$ blank characters followed by the sequence from the beginning of the vector. We use average cross entropy as the training loss function.
    The baseline solution outputs the blank category for $T+10$ time steps and then guesses a random symbol uniformly from the first eight categories. This baseline has an expected average cross entropy of $\frac{10 \log \left( 8 \right)}{T + 20}$.
    
    \begin{figure}[h]
    \centering
	\includegraphics[width=0.49\linewidth]{{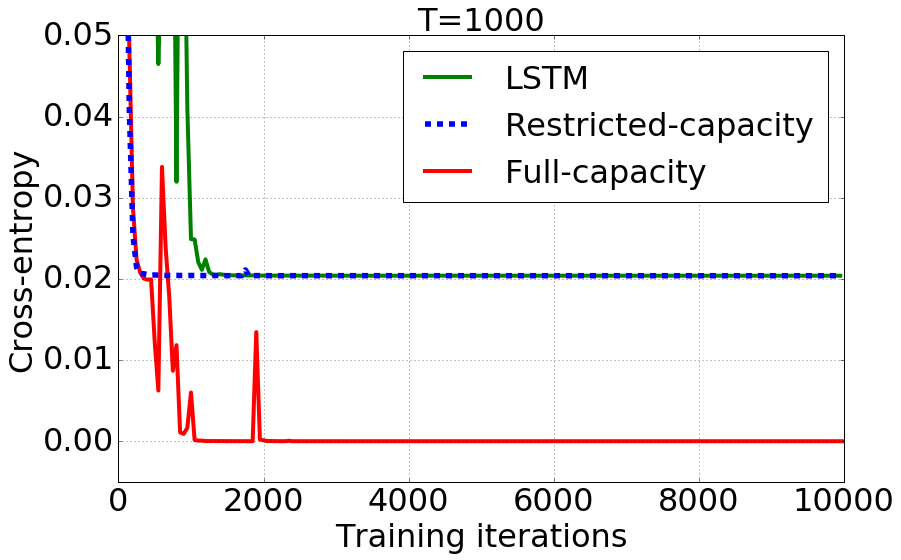}}
	\includegraphics[width=0.49\linewidth]{{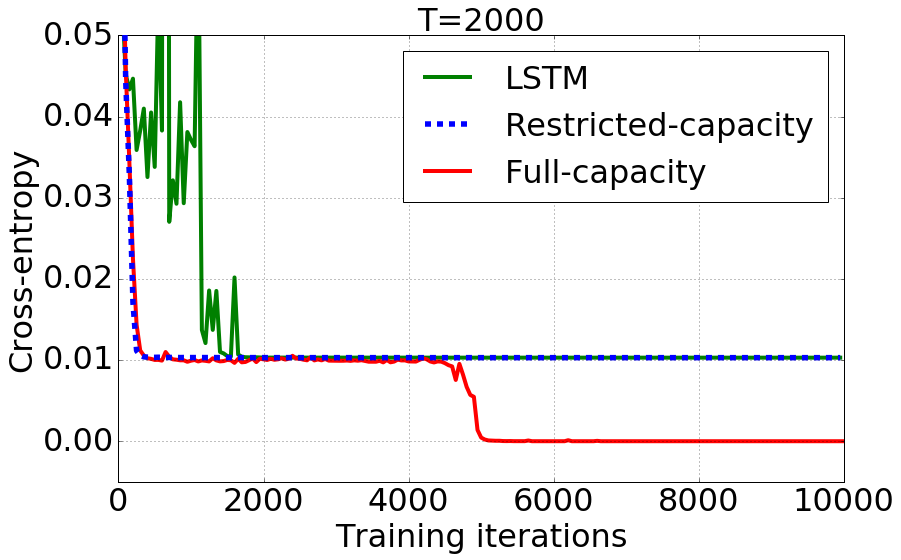}}
	\caption{Results of the copy memory problem with sequence lengths of 1000 (left) and 2000 (right). The full-capacity uRNN converges quickly to a perfect solution, while the LSTM and restricted-capacity uRNN with approximately the same number of parameters are unable to improve past the baseline naive solution.}
    \label{fig:mem}
    \end{figure}
    
    
    The full-capacity uRNN uses a hidden state size of $N=128$ with no gradient normalization. To match the number of parameters ($\approx22\mathrm{k}$), we use $N=470$ for the restricted-capacity uRNN, and $N=68$ for the LSTM.
    The training set size is 100000 and the test set size is 10000. The results of the $T=1000$ experiment can be found on the left half of Figure \ref{fig:mem}. The full-capacity uRNN converges to a solution with zero average cross entropy after about 2000 training iterations, whereas the restricted-capacity uRNN settles to the baseline solution of 0.020. The results of the $T=2000$ experiment can be found on the right half of Figure \ref{fig:mem}. The full-capacity uRNN hovers around the baseline solution for about 5000 training iterations, after which it drops down to zero average cross entropy. The restricted-capacity again settles down to the baseline solution of 0.010.
    These results demonstrate that the full-capacity uRNN is very effective for problems requiring very long memory. 
    
    \subsection{Speech data}
    
    We now apply restricted-capacity and full-capacity uRNNs to real-world speech data and compare their performance to LSTMs. 
    The main task we consider is predicting the log-magnitude of future frames of a short-time Fourier transform (STFT).
    The STFT is a commonly used feature domain for speech enhancement, and is defined as the Fourier transform of short windowed frames of the time series. In the STFT domain, a real-valued audio signal is represented as a complex-valued $F\times T$ matrix composed of $T$ frames that are each composed of $F=N_{win}/2+1$ frequency bins, where $N_{win}$ is the duration of the time-domain frame. Most speech processing algorithms use the log-magnitude of the complex STFT values and reconstruct the processed audio signal using the phase of the original observations.
    
    
    The frame prediction task is as follows: given all the log-magnitudes of STFT frames up to time $t$, predict the log-magnitude of the STFT frame at time $t+1$.
    We use the TIMIT dataset \cite{garofolo_darpa_1993}. According to common practice \cite{halberstadt1998heterogeneous}, we 
    use a training set with 3690 utterances from 462 speakers, a validation set of 400 utterances, an evaluation set of 192 utterances. Training, validation, and evaluation sets have distinct speakers. Results are reported on the evaluation set using the network parameters that perform best on the validation set in terms of the loss function over three training trials. All TIMIT audio is resampled to $8$kHz. The STFT uses a Hann analysis window of $256$ samples ($32$ milliseconds) and a window hop of $128$ samples ($16$ milliseconds). 
    
    The LSTM requires gradient clipping during optimization, while the restricted-capacity and full-capacity uRNNs do not. The hidden state dimensions $N$ of the LSTM are chosen to match the number of parameters of the full-capacity uRNN. For the restricted-capacity uRNN, we run models that match either $N$ or number of parameters. For the LSTM and restricted-capacity uRNNs, we use RMSprop \cite{tieleman_lecture_2012} with a learning rate of $0.001$, momentum $0.9$, and averaging parameter $0.1$. For the full-capacity uRNN, we also use RMSprop to optimize all network parameters, except for the recurrence matrix, for which we use stochastic gradient descent along the Stiefel manifold using the update (\ref{eq:stiefel_step}) with a fixed learning rate of $0.001$ and no gradient normalization.

    \begin{table}[h]
    \caption{Log-magnitude STFT prediction results on speech data, evaluated using objective and perceptual metrics (see text for description).}
      \label{tab:complex_pred}
      \centering
      \begin{tabular}{lccccccc}
        \toprule
        Model           & $N$   & \# parameters & \multicolumn{1}{p{0.5cm}}{\centering Valid. \\ MSE} & \multicolumn{1}{p{0.5cm}}{\centering Eval. \\ MSE} & \multicolumn{1}{p{1cm}}{\centering SegSNR \\ (dB)}   & STOI  & PESQ  \\
        \midrule
        \midrule
        LSTM            
        & 84   
        & $\approx$83k  
        %
        & 18.02 
        & 18.32   
        & 1.95	        
        & 0.77	               
        & 1.99  
        \\
        Restricted-capacity uRNN    
        & 128  
        & $\approx$67k        
        & 15.03 
        & 15.78    
        & 3.30          
        & {0.83}  
        & 2.36     
        \\
        Restricted-capacity uRNN    
        & 158  
        & $\approx$83k        
        & 15.06 
        & {\bf 14.87}    
        & 3.32          
        & {0.83} 
        & {2.33}       
        \\
        Full-capacity uRNN       
        & 128  
        & $\approx$83k       
        & {\bf 14.78}
        & { 15.24}    
        & {\bf 3.57}          
        & {\bf 0.84}  
        & {\bf 2.40}      
        \\
        \midrule
        LSTM            
        & 120   
        & $\approx$135k 
        %
        & 16.59
        & 16.98
        & 2.32
        & 0.79
        & 2.14
        \\
        Restricted-capacity uRNN    
        & 192  
        & $\approx$101k  
        %
        & 15.20
        & 15.17
        & 3.31
        & 0.83
        & 2.35
        \\
        Restricted-capacity uRNN    
        & 256  
        & $\approx$135k  
        %
        & 15.27
        & 15.63
        & 3.31
        & 0.83
        & 2.36
        \\
        Full-capacity uRNN       
        & 192  
        & $\approx$135k     
        %
        & {\bf 14.56}
        & {\bf 14.66}
        & {\bf 3.76}
        & {\bf 0.84}
        & {\bf 2.42}
        \\
        \midrule
        LSTM            
        & 158   
        & $\approx$200k   
        %
        & 15.49
        & 15.80
        & 2.92
        & 0.81
        & 2.24
        \\
        Restricted-capacity uRNN    
        & 378  
        & $\approx$200k 
        %
        & 15.78
        & 16.14
        & 3.16
        & 0.83
        & 2.35
        \\
        Full-capacity uRNN       
        & 256  
        & $\approx$200k 
        %
        & {\bf 14.41}
        & {\bf 14.45}
        & {\bf 3.75}
        & {\bf 0.84}
        & {\bf 2.38}
        \\
        \bottomrule
      \end{tabular}
    \end{table}

    \begin{figure}[ht]
        \centering
        \includegraphics[width=0.95\linewidth]{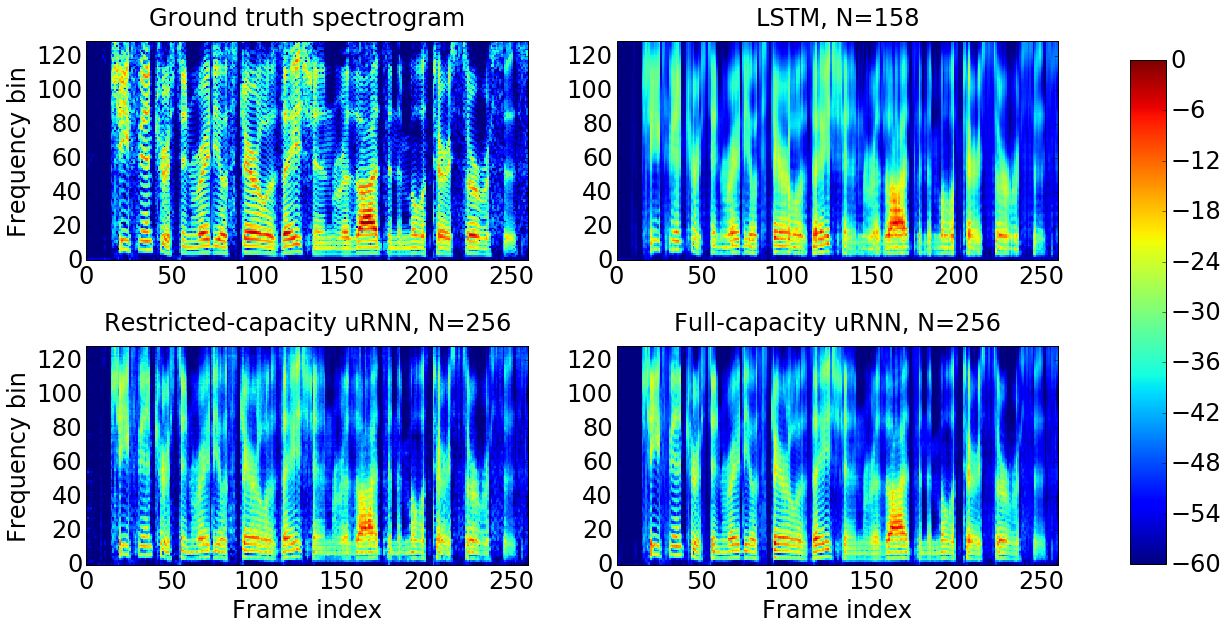}
        \vspace{-10pt}
        \caption{Ground truth and one-frame-ahead predictions of a spectrogram for an example utterance. For each model, hidden state dimension $N$ is chosen for the best validation MSE. Notice that the full-capacity uRNN achieves the best detail in its predictions.
        }\label{fig:sgrams}
        \vspace{-7.5pt}
    \end{figure}
    
    Results are shown in Table \ref{tab:complex_pred}, and Figure \ref{fig:sgrams} shows example predictions of the three types of networks. Results in Table \ref{tab:complex_pred} are given in terms of the mean-squared error (MSE) loss function and several metrics computed on the time-domain signals, which are reconstructed from the predicted log-magnitude and the original phase of the STFT. These time-domain metrics are segmental signal-to-noise ratio (SegSNR), short-time objective intelligibility (STOI), and perceptual evaluation of speech quality (PESQ). SegSNR, computed using \cite{brookes_voicebox:_2002}, uses a voice activity detector to avoid measuring SNR in silent frames. 
    STOI is designed to correlate well with human intelligibility of speech, and takes on values between 0 and 1, with a higher score indicating higher intelligibility \cite{taal_algorithm_2011}. PESQ is the ITU-T standard for telephone voice quality testing \cite{rix_perceptual_2001,itu-t_p.862_2000}, and is a popular perceptual quality metric for speech enhancement \cite{loizou_speech_2007}. PESQ ranges from 1 (bad quality) to 4.5 (no distortion).

    Note that full-capacity uRNNs generally perform better than restricted-capacity uRNNs with the same 
    number of parameters,
    and both types of uRNN significantly outperform LSTMs. 

    \subsection{Pixel-by-pixel MNIST}
    
    As another challenging long-term memory task with natural data, we test the performance of LSTMs and uRNNs on pixel-by-pixel MNIST and permuted pixel-by-pixel MNIST, first proposed by \cite{le_simple_2015} and used by \cite{arjovsky_unitary_2016} to test restricted-capacity uRNNs. For permuted pixel-by-pixel MNIST, the pixels are shuffled, thereby creating some non-local dependencies between pixels in an image. Since the MNIST images are $28\times28$ pixels, resulting pixel-by-pixel sequences are $T=784$ elements long. We use 5000 of the 60000 training examples as a validation set to perform early stopping with a patience of 5. The loss function is cross-entropy. Weights with the best validation loss are used to process the evaluation set. The full-capacity uRNN uses RMSprop-style gradient normalization.
    
        \begin{table}[!ht]
    \caption{Results for unpermuted and permuted pixel-by-pixel MNIST. Classification accuracies are reported for trained model weights that achieve the best validation loss.}\vspace{-.1cm}
      \label{tab:mnist}
      \centering

      \begin{tabular}{clcccc}
        \toprule
        &Model           & $N$   & \# parameters & Validation accurary & Evaluation accuracy \\
        \midrule
        \midrule
        \multirow{ 5}{*}{\rotatebox{90}{Unpermuted}}
        &LSTM                        & 128 & $\approx$~68k   & 98.1    & 97.8	\\
        &LSTM                        & 256 & $\approx$270k  & {\bf 98.5}      & {\bf 98.2}         \\
        &Restricted-capacity uRNN    & 512  & $\approx$~16k   & 97.9    & 97.5   \\
        &Full-capacity uRNN          & 116  & $\approx$~16k   & 92.7    & 92.8   \\
        &Full-capacity uRNN          & 512  & $\approx$270k  & {97.5}    & {96.9}      \\
        \midrule
        \multirow{ 5}{*}{\rotatebox{90}{Permuted}}
        &LSTM                        & 128 & $\approx$~68k   & 91.7    & 91.3	\\
        &LSTM                        & 256 & $\approx$270k  & 92.1      & 91.7         \\
        &Restricted-capacity uRNN    & 512  & $\approx$~16k   & 94.2    & 93.3   \\
        &Full-capacity uRNN          & 116  & $\approx$~16k   & 92.2    & 92.1   \\
        &Full-capacity uRNN          & 512  & $\approx$270k  & {\bf 94.7}    & {\bf 94.1}      \\
        \bottomrule
      \end{tabular}
    \end{table}

\begin{figure}[t]
        \centering
        \includegraphics[width=0.83\linewidth]{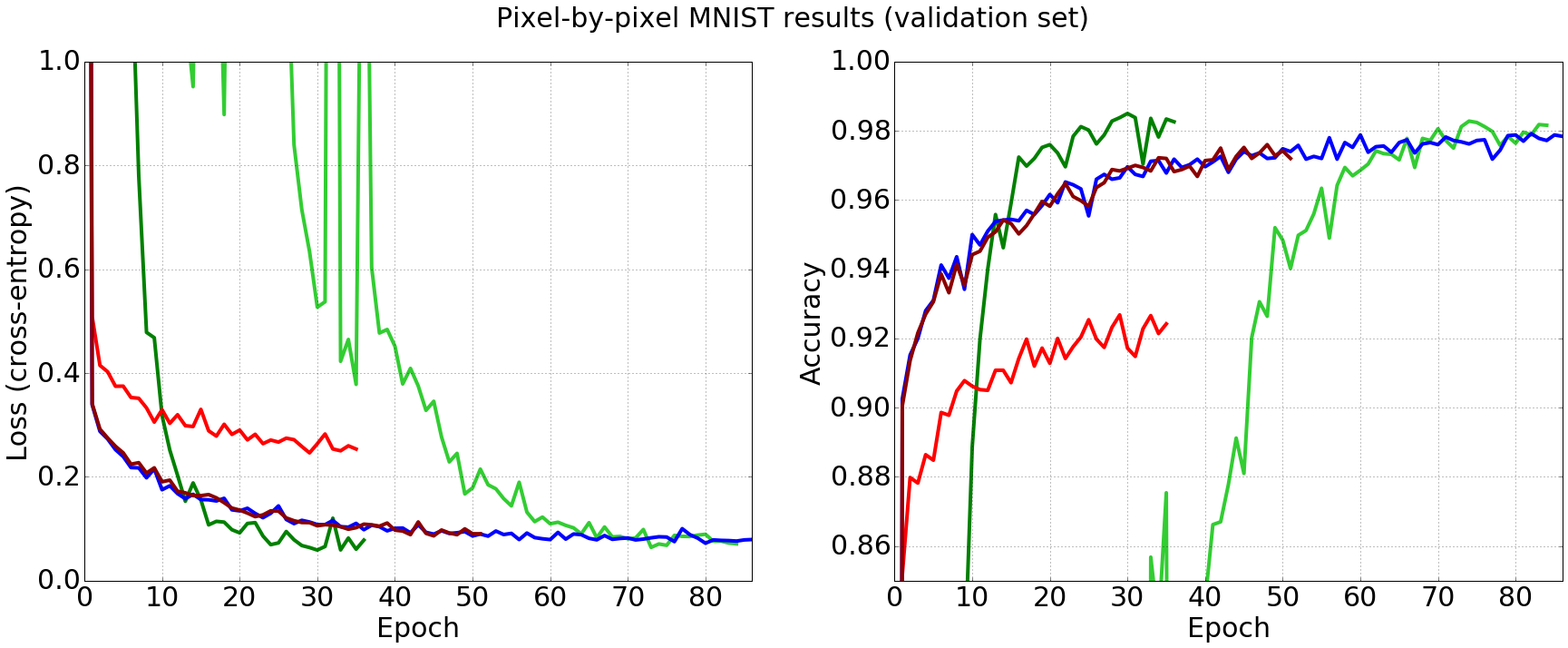}
         \includegraphics[width=0.83\linewidth]{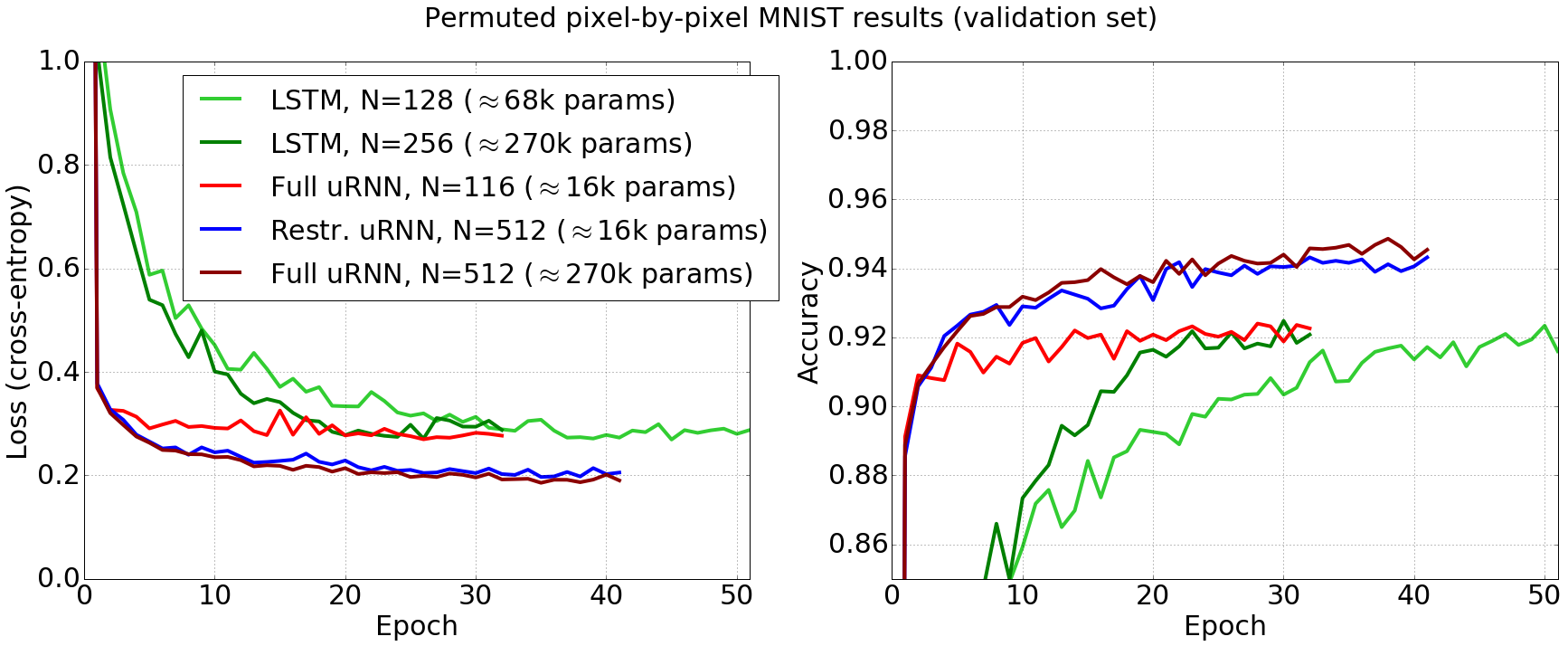}
        \vspace{-8pt}
        \caption{Learning curves for unpermuted pixel-by-pixel MNIST (top panel) and permuted pixel-by-pixel MNIST (bottom panel).}
        \vspace{-11pt}\label{fig:mnist}        
    \end{figure}

    Learning curves are shown in Figure \ref{fig:mnist}, and a summary of classification accuracies is shown in Table \ref{tab:mnist}. For the unpermuted task, the LSTM with $N=256$ achieves the best evaluation accuracy of $98.2\%$. For the permuted task, the full-capacity uRNN with $N=512$ achieves the best evaluation accuracy of $94.1\%$, which is state-of-the-art on this task. Both uRNNs outperform LSTMs on the permuted case, achieving their best performance after fewer traing epochs and using an equal or lesser number of trainable parameters. This performance difference suggests that LSTMs are only able to model local dependencies, while uRNNs have superior long-term memory capabilities. Despite not representing all unitary matrices, the restricted-capacity uRNN with $N=512$ still achieves impressive test accuracy of $93.3\%$ with only $1/16$ of the trainable parameters, outperforming the full-capacity uRNN with $N=116$ that matches number of parameters. 
    This result suggests that further exploration into the potential trade-off between hidden state dimension $N$ and capacity of unitary parameterizations is necessary.

\vspace{-.1cm}
    \section{Conclusion}
 \vspace{-.1cm}
    
    Unitary recurrent matrices prove to be an effective means of addressing the vanishing and exploding gradient problems. 
    We provided a theoretical argument to quantify the capacity of constrained unitary matrices.
    We also described a method for directly optimizing a full-capacity unitary matrix by constraining the gradient to lie in the differentiable 
    manifold of unitary matrices. The effect of restricting the capacity of the unitary weight matrix was tested on system identification and memory tasks, in which full-capacity unitary recurrent neural networks (uRNNs) outperformed restricted-capacity uRNNs from \cite{arjovsky_unitary_2016} as well as LSTMs.
    Full-capacity uRNNs also outperformed restricted-capacity uRNNs on log-magnitude STFT prediction of natural speech signals and classification of permuted pixel-by-pixel images of handwritten digits, and both types of uRNN significantly outperformed LSTMs.
    In future work, we plan to explore more general forms of restricted-capacity unitary matrices, including constructions based on products of elementary unitary matrices such as Householder operators or Givens operators.  

{\bf Acknowledgments:}
We thank an anonymous reviewer for suggesting improvements to our proof in Section \ref{sec:capacity} and Vamsi Potluru for helpful discussions. 
Scott Wisdom and Thomas Powers were funded by U.S. ONR contract number N00014-12-G-0078, delivery orders 13 and 24. Les Atlas was funded by U.S. ARO grant W911NF-15-1-0450. 


\vfill\pagebreak
\section*{References}
\vspace{.1cm}

\small

\begingroup
\renewcommand{\section}[2]{} 
\renewcommand*{\bibfont}{\footnotesize} 
\setlength{\bibsep}{0.5ex}
\bibliographystyle{unsrt_abbrv}
\bibliography{nips2016}
\endgroup

\end{document}